%% file: main.tex
\documentclass[conference]{IEEEtran}
\IEEEoverridecommandlockouts
\usepackage{cite}
\usepackage{amsmath,amssymb,amsfonts}
\usepackage{algorithmic}
\usepackage{graphicx}
\usepackage{textcomp}
\usepackage{xcolor}
\usepackage{multirow}
\usepackage{float}
\usepackage{acronym}
\def\BibTeX{{\rm B\kern-.05em{\sc i\kern-.025em b}\kern-.08em
    T\kern-.1667em\lower.7ex\hbox{E}\kern-.125emX}}

\usepackage{cleveref}
\crefname{supp}{supplement}{Supplements}
\crefname{app}{appendix}{Appendices}
\crefformat{equation}{Eq~(#2#1#3)}
\crefformat{figure}{Fig.\,#2#1#3}
\crefformat{section}{\,#2#1#3}
\crefformat{appendix}{App.\,#2#1#3}
\crefformat{table}{Tab.\,#2#1#3}

\acrodef{DL}[DL]{Deep Learning}
\acrodef{SNN}[SNN]{Spiking Neural Network}
\acrodef{ANN}[ANN]{Artificial Neural Network}

\usepackage{changes} 
\definecolor{HB}{RGB}{0,200,0}
\definecolor{AO}{RGB}{200,0,200}
\definecolor{IH}{RGB}{200,150,150}
\definecolor{YB}{RGB}{0,150,255}
\definecolor{REV}{RGB}{220,0,0}
\definechangesauthor[color=TT]{TT}
\definechangesauthor[color=DW]{DW}
\definechangesauthor[color=MD]{MD}
\definechangesauthor[color=YB]{YB}
\definechangesauthor[color=REV]{REV}

\makeatletter 
\newcommand{\linebreakand}{%
  \end{@IEEEauthorhalign}
  \hfill\mbox{}\par
  \mbox{}\hfill\begin{@IEEEauthorhalign}
}
\makeatother 

\begin{document}

\title{Skip Connections in Spiking Neural Networks: An Analysis of Their Effect on Network Training
\thanks{* Completed during a School of AI Algiers reading session at Ecole Nationale Supérieure d'Informatique, Algiers, Algeria.}}

\author{
\IEEEauthorblockN{Hadjer Benmeziane*}
\IEEEauthorblockA{Univ. Polytechnique \\ Hauts-de-France,  \\ Valenciennes, France \\ 
hadjer.benmeziane@uphf.fr}
\and
\IEEEauthorblockN{Amine Ziad Ounnoughene*}
\IEEEauthorblockA{Belmihoub Abd El Rahmane \\ High School \\ Bordj bou arreidj, Algeria \\ amine.ziad.ounnoughene@gmail.com
}
\linebreakand
\IEEEauthorblockN{Imane Hamzaoui*}
\IEEEauthorblockA{Ecole Nationale\\ Supérieure d'Informatique \\ Algiers,Algeria \\
ji\_hamzaoui@esi.dz}
\and
\IEEEauthorblockN{Younes Bouhadjar*}
\IEEEauthorblockA{Peter Gr\"unberg Institute (PGI-7,15), \\ Forschungszentrum J\"ulich, Germany \\
y.bouhadjar@fz-juelich.de}
}

\maketitle


\begin{abstract}
Spiking neural networks (SNNs) have gained attention as a promising alternative to traditional artificial neural networks (ANNs) due to their potential for energy efficiency and their ability to model spiking behavior in biological systems. 
However, the training of SNNs is still a challenging problem, and new techniques are needed to improve their performance. In this paper, we study the impact of skip connections on SNNs and propose a hyperparameter optimization technique that adapts models from ANN to SNN. We demonstrate that optimizing the position, type, and number of skip connections can significantly improve the accuracy and efficiency of SNNs by enabling faster convergence and increasing information flow through the network. Our results show an average +8\% accuracy increase on CIFAR-10-DVS and DVS128 Gesture datasets adaptation of multiple state-of-the-art models. 
\end{abstract}


\input{sections/introduction}

\input{sections/related_works}

\input{sections/method}

\input{sections/experiments}

\section{Conclusion and Future directions}
This paper presents novel insights into the design and training of spiking neural networks (SNNs) and highlights the potential of skip connections as a promising tool for advancing SNN research. Our study evaluated both densenet-like and addition-type skip connections and found that both improved accuracy, with densenet-like connections being more energy-efficient by slightly increasing the firing rate. Our comprehensive hyperparameter optimization process led to the discovery of the optimal ANN to SNN adaptation, resulting in an average accuracy improvement of 8\% within approximately 5 minutes. 
These results demonstrate the significance of skip connections in designing and training SNNs and pave the way for further research in this field.
 In future work, we plan to further improve the performance of SNNs by incorporating backward connections into our hyperparameter optimization. Additionally, exploring the split and connectivity between ANN/SNN processing on edge devices and the cloud is another promising avenue for future research. The integration of these innovations could lead to more practical applications of SNNs in real-world scenarios.
\vspace{-0.25cm}
\bibliographystyle{IEEEtran}
\bibliography{references}

\end{document}

%% file: sections/introduction.tex
\section{Introduction}

While \ac{DL} has become a foundational technology to many applications, including autonomous driving, and medical imagining segmentation. Its inference is still energy-, resource- and time-expensive.

An ongoing challenge in \ac{DL} research is to devise energy-efficient inference algorithms and hardware platforms. Taking inspiration from the brain, two intertwined solutions are investigated: (1) designing neuromorphic hardware, in which memory and computations reside within the same node and (2) building bio-inspired networks such as \ac{SNN}, which can be more energy efficient than nonspiking neural networks. 
In this paper, we refer to this latter as \ac{ANN}. 

Although \ac{ANN} are brain-inspired, there are fundamental differences in their computations and learning mechanism compared to the brain. 
In the brain, neurons communicate with each other via sequences of an action potential, also known as \textit{spikes}. 
An action potential travels along the axon in a neuron and activates synapses. 
The synapses release neurotransmitters that arrive at the postsynaptic neuron. 
Here, the action potential rises with each incoming pulse of neurotransmitters. 
If the action potential reaches a certain threshold the postsynaptic neuron fires a spike itself. 
These individual spikes are sparse in time and space, which is the main reason behind the energy efficiency of biological neuronal networks, i.e., SNNs.
In addition, the information in SNN is conveyed by spike times and thus can have a large capacity for encoding and representing the input data at the edge.

However, \ac{SNN} architecture design and training are still in their early phases. An important scientific question is to what extent architecture characteristics (e.g., operations, skip connections) are compliant with the spatial and temporal constraints in \ac{SNN}. 
To answer this question, this paper presents the following contribution: 

\begin{itemize}
    \item We investigate the relationship between the number of skip connections and accuracy drop that comes from standard architectures, including denseNet121~\cite{HuangLW16a}, resnet18~\cite{HeZRS15}, and mobilenetv2~\cite{mobilenet-v2} converted to \ac{SNN}. To the best of our knowledge, this is the first work introducing a dense spiking neural network.
    
    \item We propose an adaptation hyperparameter tuning algorithm that selects the best number of skip connections to optimize the trade-off between accuracy drop and energy efficiency. 
\end{itemize}

%% file: sections/related_works.tex
\begin{figure*}[ht!]
    \centering
    \includegraphics[width=\textwidth]{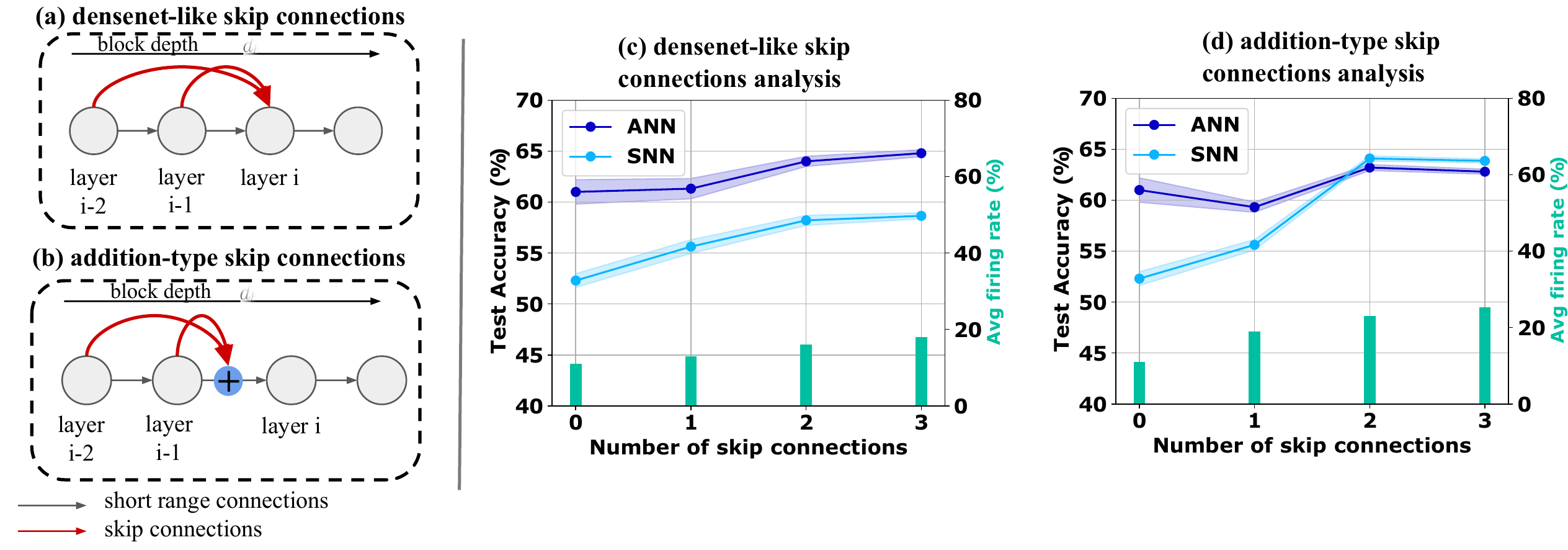}
    \caption{Left: Commonly used skip connections in neural networks. Right: Skip connections investigation results.}
    \label{fig:skip_types}
\end{figure*}
\section{Related Works}
\ac{ANN}s use gradient descent to optimize the weights during training, but this has proven challenging in \ac{SNN} due to their non-differentiable nonlinear spikes \cite{Neftci19_51}. 
A solution to this problem was to introduce surrogate derivatives during the backpropagation (BP) operation, which replaces the spiking activations with a smooth function \cite{Zenke21_899}.
While this has been successful at improving the accuracy of SNNs in solving certain problems, their performance often falls short compared to that of ANNs.
The reason for this discrepancy is twofold: 1) the surrogate gradient approaches only provide an approximation of the gradients
and 2) the unfolding of SNNs in time to perform the backpropagation BP leads to the vanishing gradient problem, similar to the problem faced in vanilla Recurrent Neural Networks (RNNs).
A study such as in \cite{Wunderlich21_12829} provides a method that computes exact gradients for arbitrary SNN architecture, but their applications were limited to relatively simple datasets, e.g., MNIST.
The incorporation of local losses in space \cite{Kaiser20_424} and time \cite{Zenke18_1541} has shown promising results in circumventing the vanishing gradient problem, but as these methods only roughly approximate the gradients, they lead to less competitive performance.
To further improve the learning in SNNs, other studies took a different approach by introducing novel architectures and ingredients. For example, the work in \cite{Kim21_1638} showed that batch normalization through time could effectively train deep SNN.
Another example is presented in
\cite{DBLP:conf/eccv/KimLPVP22}, 
where the authors implemented a dedicated neural architecture search (NAS) to find the best architecture within common NAS benchmarks, such as NAS-Bench-201. While their methodology is promising, we found that adapting ANN standard architectures such as resnet18, or densenet, yields better accuracies in less search time.
In our study, we adapt standard architectures and explore the effect of skip connections on learning in SNNs.

%% file: sections/method.tex
\section{Proposed Methodology}
In this section, we first describe the skip connection investigation. The investigation aims at understanding the importance of skip connections in \ac{SNN} and motivating the use of hyperparameter optimization on the number of skip connections. We then present the general steps of our hyperparameters optimization strategy. 

\subsection{Skip connections in SNN}
Common state-of-the-art topologies comprise a small repeated set of layers, called \textit{blocks}. The blocks are connected with a single sequential connection. The block's topology is described as a directed acyclic graph (DAG). Each vertex corresponds to a layer such as convolution, attention, or fully connected. The number of vertices in the graph is referred to as the depth of the block, $d_b$. An adjacency matrix represents the connections between vertices. 


Skip connections are commonly used inside the blocks to increase overall accuracy. Their main goal is to overcome the vanishing gradient problem that may prevent layers from training. 
There are two types of skip connections in the literature, as shown in figure~\ref{fig:skip_types} (a) and (b). 

\begin{itemize}
    \item Densenet-like Skip Connections (DSC)~\cite{HuangLW16a} concatenates previous layers' outputs, $l_{i-n}$ where $0<n<i$, as the input to the next layer. A direct mathematical relation is then created between the weights of layer $l_{i-n}$ and the output of layer $l_i$, which enhances backward gradient computation. However, adding these connections enlarges the input tensor which augments the number of multiply-accumulates operations (MAC). 

    \item Addition-type Skip Connections (ASC)~\cite{HeZRS15} perform element-wise summation of the outputs of previous layers, $l_{i-n}$ where $0<n<i$. The result is the input to the current layer $l_i$. This type is usually used in resnet-like architectures.  
\end{itemize}
To study the topological properties, we do not use the original architectures, such as DenseNets which contains all-to-all connections. 
Instead, we consider a generalized version where we vary the number of skip connections by randomly selecting only some channels for concatenation.
\begin{figure*}[t!]
    \centering
    \includegraphics[trim=0 1cm 0 0, clip,width=\textwidth]{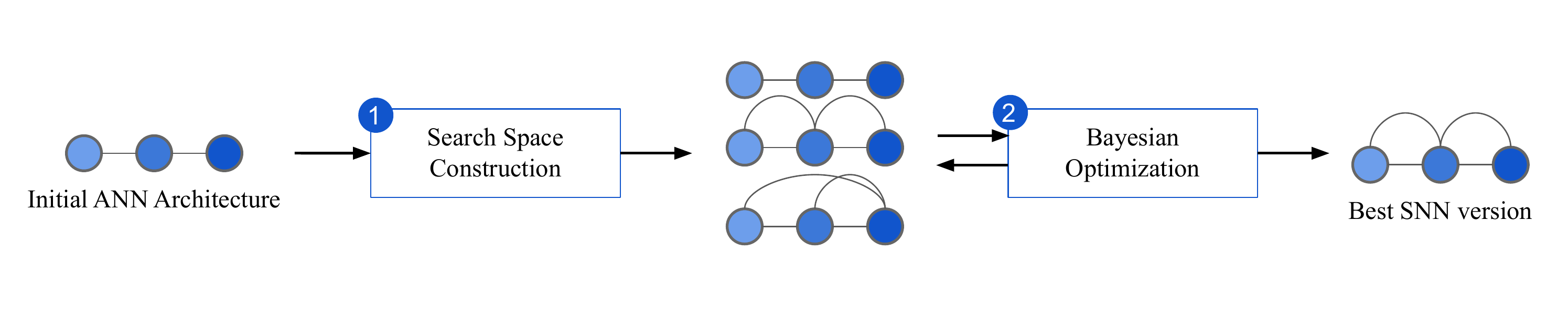}
    \caption{Overview of the hyperparameters optimization process.}
    \label{fig:overview}
    \vspace{-0.25cm}
\end{figure*}

We define $n_{\text{skip},i}$ as the number of skip connections coming to layer $i$. To analyze the skip connection effect, we first build a single-block architecture, with 4 convolution layers inside the block. Figure~\ref{fig:skip_types} (right) shows the results of varying the $n_\text{skip}$ and using both DSC and ASC types of skip connections. If $n_\text{skip}$ is greater than the number of previous layers, we use the number of previous layers instead. For example, the second layer can only have $n_\text{skip} = 1$, and the fourth layer can have up to $n_\text{skip} = 3$. The hyperparameters and setup are described in section~\ref{sec:setup}. We show the test accuracy on CIFAR-DVS~\cite{Li2017} as well as the average firing rate for the SNN models. The firing rate refers to the rate at which a block generates output signals, which are typically referred to as spikes. 

From figure~\ref{fig:skip_types} (right), we draw the following conclusions: 
\begin{itemize}
    \item Overall, increasing the number of skip connections, regardless of their type, enhances the model's accuracy and decreases the drop from the corresponding ANN version. 
    \item Adding DSC slightly increases the average firing rate compared to ASC. The firing rate of the original architecture, i.e., $n_\text{skip} = 0$, is low (11\%). The summation operation adds up the spikes of multiple inputs, which increases the overall firing rate of the network, while the concatenation operation combines the input signals into a single signal, which results in a lower overall firing rate but yields an increasing number of MACs.
    The choice of connection type between layers is critical.
   A lower firing rate and fewer operations can lead to energy efficiency and reduced computational requirements while offering a decent increase in accuracy.
\end{itemize}

\subsection{Skip Connections Optimization}
\label{sec:skip_connections_optimization}

We define the skip connections in the adjacency matrix of each block $A_b$, where $b$ is the index of the block in the overall topology. Each element of the adjacency matrix is defined in equation~\ref{eq:aij}. Note that the adjacency matrix does not include any backward connections. 

\begin {equation}
    a_{ij} = \left\{
\begin{array}{ll}
 0 & \text{if no connection between i and j } \\
 1 & \text{DSC connection between i and j } \\
 2 & \text{ASC connection between i and j } 
\end{array}
\right.
\label{eq:aij}
\end{equation}

Figure~\ref{fig:overview} shows the overall hyperparameter optimization strategy used to adapt a given ANN such as densenet~\cite{HuangLW16a}, resnet~\cite{HeZRS15}, or mobilenetv2~\cite{mobilenet-v2} to SNN. Given an initial ANN topology, denoted as $\alpha$, the optimization aims at finding the right number, position, and type of skip connections that minimize the drop between ANN accuracy and its SNN counterpart. The overall optimization process comprises two steps: (1) we begin by constructing the search space of all possible adjacency matrices. Each block is extracted from the given topology and the number of layers in each block as well as the initial adjacency matrices are defined. (2) We use bayesian optimization (BO) to optimize the accuracy drop. 

Formally, BO seeks to compute $A^* = argmin_{A \in \Lambda}(A)$, where $\Lambda$ is the set of all possible adjacency matrices and $f(A)$ denotes the accuracy drop between the topology obtained with the adjacency matrix $A$ and accuracy of $\alpha$. 
Over a sequence of iterations, the results from all previous iterations are used to model
the topology of $\{f(A)\}_{A \in \Lambda}$ using the posterior distribution of the model. The next architecture is then chosen by optimizing an acquisition function. Two design decisions need to be taken in the case of BO: (1) The prior model, and (2) The acquisition function. 

\begin{itemize}
    \item \textbf{The Prior:} defines the probability distribution over the objective function. We use a gaussian process~\cite{DBLP:conf/aistats/Song0Y19} (GP) to model this distribution. GP is a valuable surrogate model for such an expensive training objective. We do not use a predictor as it would require creating a dataset for each given topology. 
    \item \textbf{The acquisition function:} defines how we select the next point to sample, given a conditional distribution over the values of the objective given by the prior. The most common acquisition functions used in literature are expected improvement (EI), probability of improvement (PI), and upper confidence bound (UCB)~\cite{DBLP:journals/jmlr/Auer02}. The latter is used in our search strategy. The UCB algorithm enables us to balance exploration and exploitation. It shifts from concentrating on exploration, choosing the least preferred actions, to focusing on exploitation.
\end{itemize}

These functions balance exploration with exploitation during the search. 
The chosen architecture is then trained and used to update GP. Evaluating $f(A)$ in each iteration is the bottleneck of BO since we need to train the model. Because we optimize the skip connections, we can use previously trained weights and share them among all possible topologies $\{f(A)\}_{A \in \Lambda}$. We only fine-tune the networks for $n$ epochs to account for the removal or addition of skip connections. Besides, we use parallel BO, i.e., our strategy outputs $k$ architectures to train in each iteration, so that the $k$ architectures can be trained in parallel.

%% file: sections/experiments.tex
\begin{table*}[ht!]
\centering
\caption{Comparison results on CIFAR-10, CIFAR-10-DVS, and DVS128 Gesture.}
\begin{tabular}{l|l|c|c|c|c|c}
\hline
dataset                         & model       & \multicolumn{1}{l|}{\begin{tabular}[c]{@{}l@{}}ANN \\ accuracy (\%)\end{tabular}} & \multicolumn{1}{l|}{\begin{tabular}[c]{@{}l@{}}SNN \\ accuracy (\%)\end{tabular}} & \multicolumn{1}{l|}{\begin{tabular}[c]{@{}l@{}}Our Optimized \\ SNN Accuracy (\%)\end{tabular}} & \multicolumn{1}{l|}{\begin{tabular}[c]{@{}l@{}}SNN avg \\ firing rate\end{tabular}} & \multicolumn{1}{l}{\begin{tabular}[c]{@{}l@{}}Optimized avg \\ firing rate\end{tabular}} \\ \hline
\multirow{3}{*}{CIFAR-10}       & resnet18    & 93.02 (+/- 1.1\%)                                                                 & 81.3 (+/- 1.4\%)                                                                  & \textbf{90.34 (+/- 0.2\%)}                                                                      & 15.6\%                                                                              & \textbf{22.3\%}                                                                           \\ \cline{2-7} 
                                & densenet121 & 95.03 (+/- 0.3\%)                                                                 & 89.23 (+/- 2.1\%)                                                                 & \textbf{95.45 (+/- 0.05\%)}                                                                     & 17.4\%                                                                              & \textbf{21\%}                                                                             \\ \cline{2-7} 
                                & mobilenetv2 & 94.43 (+/- 0.8\%)                                                                 & 73.14 (+/- 1.3\%)                                                                 & \textbf{92.31 (+/- 0.1\%)}                                                                      & 7.5\%                                                                               & \textbf{16.83\%}                                                                          \\ \hline\hline
\multirow{3}{*}{CIFAR-10-DVS}   & resnet18    & -                                                                                 & 60.43 (+/- 1.5\%)                                                                 & \textbf{65.23 (+/- 0.1 \%)}                                                                     & 11.2\%                                                                              & \textbf{17.32\%}                                                                          \\ \cline{2-7} 
                                & densenet121 & -                                                                                 & 64.34 (+/- 1.2\%)                                                                 & \textbf{75.34 (+/- 0.14\%)}                                                                     & 14.23\%                                                                             & \textbf{15.3\%}                                                                           \\ \cline{2-7} 
                                & mobilenetv2 & -                                                                                 & 62.7 (+/- 2.3\%)                                                                  & \textbf{74.32 (+/- 0.2\%)}                                                                      & 8.67\%                                                                              & \textbf{16.43\%}                                                                          \\ \hline\hline
\multirow{3}{*}{DVS128 Gesture} & resnet18    & -                                                                                 & 90.91 (+/- 0.6\%)                                                                 & \textbf{95.43 (+/- 0.15\%)}                                                                     & 10.32\%                                                                             & \textbf{13.7\%}                                                                           \\ \cline{2-7} 
                                & densenet121 & -                                                                                 & 91.2 (+/- 1.3\%)                                                                  & \textbf{93.21 (+/- 0.03\%)}                                                                     & 9.6\%                                                                               & \textbf{15.4\%}                                                                           \\ \cline{2-7} 
                                & mobilenetv2 & -                                                                                 & 70.32 (+/- 1.12\%)                                                                & \textbf{94.5 (+/- 0.11\%)}                                                                      & 6.4\%                                                                               & \textbf{11.5\%}                                                                           \\ \hline
\end{tabular}

\label{tab:results}
\end{table*}

\section{Experiments}
\label{sec:setup}
In this section, we present the results of the effects of skip connections on the performance of SNN.
The experiments were conducted using the snnTorch library~\cite{eshraghian2021training}. We evaluated the optimization on three datasets: CIFAR-10~\cite{krizhevsky2009learning}, CIFAR-10-DVS~\cite{Li2017}, and DVS128 Gesture~\cite{DVS_gesture_8100264}. 
\par
For CIFAR-10, we trained the model on 50k images split into 10 classes, with 5k each for validation and testing. We used an SGD optimizer with a learning rate of 0.01 and a momentum of 0.9. The employed number of steps is equal to 25 and the maximum number of epochs is equal to 200.
For CIFAR-10-DVS, the dataset contains 10k images recorded by a DVS128 sensor, split into 90\% training and 10\% test, with the training further divided into 80\% new training and 20\% validation. We train the models for 100 epochs with a learning rate set to 0.025, SGD optimizer, and momentum to 0.9.
For DVS128 Gesture, event data was recorded from 29 subjects performing 11 hand gestures and preprocessed into 4k sequences for training, 500 for validation, and 500 for testing. The training was performed for 200 epochs with a learning rate of 0.01 and Adam optimizer.

\subsection{Overall results}
Table~\ref{tab:results} shows the overall results of optimizing the skip connections on three state-of-the-art architectures resnet18~\cite{HeZRS15}, densenet121~\cite{HuangLW16a}, and mobilenetv2~\cite{mobilenet-v2}. 
Our optimized architectures consistently perform better than the original SNN version. On average, we achieve 
+11.3\%, +9.3\%, and +10.2\% accuracy on CIFAR-10, CIFAR-10-DVS, and DVS128 Gesture respectively. Note that given the non-static images in CIFAR-10-DVS and DVS128 Gesture, conventional ANN is omitted for such type of data. 

\subsection{Comparison with random Search}
\begin{figure}
    \centering
    \includegraphics[width=0.45\textwidth]{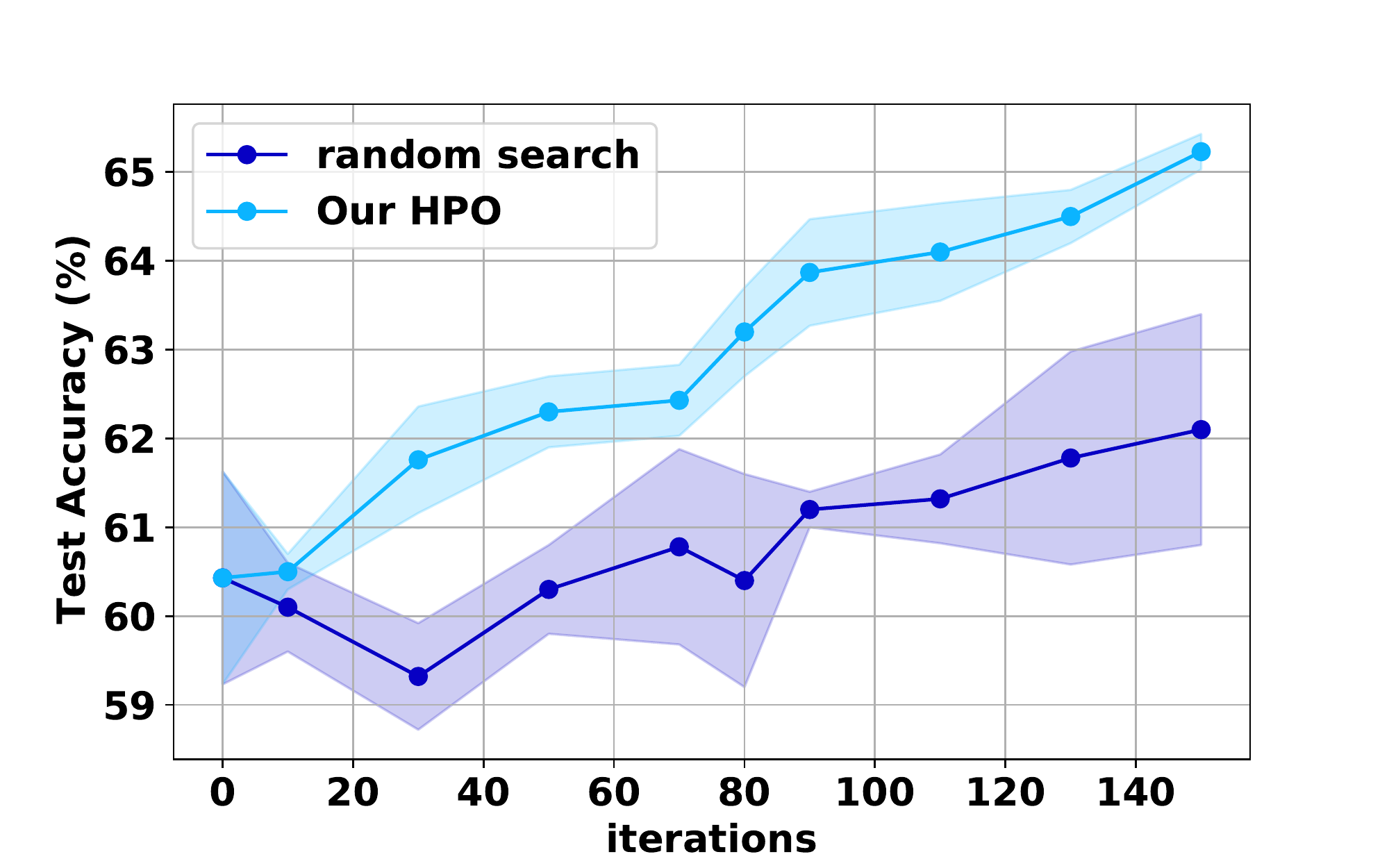}
    \caption{Comparison with random search. The shaded area provides the standard deviation obtained after 5 runs.}
    \vspace{-0.5cm}
    \label{fig:rs}
\end{figure}

The efficacy of our hyperparameter optimization is compared to random search in figure~\ref{fig:rs}. We implement a random search (RS) over the constructed search space of adjacency matrices. The search builds the architecture based on a randomly sampled adjacency matrix without replacement. Then RS trains the architecture from scratch which requires a massive computing budget. With fewer iterations, our method achieves better performance for optimizing multiple models with fewer iterations. Besides, our hyperparameter optimization is more stable and consistently provides closer solutions, which ensures finding the right architectures within a single run of search.

